\documentclass[runningheads]{llncs}
\usepackage[T1]{fontenc}
\usepackage{graphicx}
\usepackage{booktabs}
\usepackage{float}
\usepackage[most]{tcolorbox}
\usepackage{adjustbox}
\usepackage{tabularx}
\usepackage{booktabs}
\usepackage{array}
\newcolumntype{L}[1]{>{\raggedright\arraybackslash}p{#1}} 
\usepackage{soul}
\usetikzlibrary{positioning}
\usepackage{hyperref}
\usepackage{amsfonts} 
\usepackage{siunitx}
\usepackage{enumitem}
\setlist{nosep}
\usepackage[numbers]{natbib}

\usepackage{color}

\begin{document}
\title{Mitigating Gender Bias in English to Romanian Machine Translation}
\titlerunning{Gender Bias in Romanian MT}
%
\author{Ioana Grigore\orcidID{0009-0006-7370-8159} \and
Sergiu Nisioi\orcidID{0000-0003-2247-4488} }
\authorrunning{Grigore and Nisioi, 2026}
%
\institute{Human Language Technologies Research Center \\ Faculty of Mathematics and Computer Science \\ University of Bucharest \\
\email{ioanaagrigore28@gmail.com, sergiu.nisioi@unibuc.ro}}
%
\maketitle              
\begin{abstract}
Machine translation (MT) systems often fail to correctly translate gender, especially when converting from a gender-neutral language like English to a gendered target language such as Romanian. This bias results in translations that default to masculine forms or reinforce gender stereotypes. We propose a hybrid pipeline to mitigate this issue by combining large language model (LLM)-based gender classification with neural machine translation (NMT). Our system uses a fine-tuned LLM to detect the intended gender of target words in English sentences and insert inline gender hint tags. These tagged sentences are then passed to a Transformer model fine-tuned to generate morphologically correct Romanian translations. To support this, we introduce three novel datasets for gender disambiguation and translation. Our approach improves gender accuracy on the WinoMT and WinoGender benchmarks by over 40 percentage points compared to a baseline MT system. This is the first method to explicitly address and evaluate gender bias in English–Romanian MT using both LLM inference and tag-aware translation.

\end{abstract}
\section{Introduction}

Gender bias in machine translation (MT) remains a well-documented challenge, particularly when translating from languages like English—where gender is often implicit—to target languages such as Romanian, which require explicit grammatical gender agreement. Most neural MT systems tend to default to masculine forms or fail to resolve gender correctly from context, resulting in biased or grammatically incorrect outputs. This issue is especially prominent in translations involving professions, roles, or named entities referring to people.

In this work, we introduce novel datasets and a hybrid pipeline that combines large language model (LLM)-based gender inference with a fine-tuned neural machine translation (NMT) model. Specifically, we use a fine-tuned LLaMA model to classify the gender of target words in an English sentence and insert explicit inline gender hint tags (e.g., \texttt{<tgF>teacher</tgF>}). A Transformer model is then trained to recognize and act on these tags, producing gender-aware Romanian translations.

To support this system, we release\footnote{All data is released under CC BY-NC 4.0 license at \url{https://github.com/Ioannnnna/EnRoGend}.} novel, high-quality corpora that support (i) gender-aware English classification and (ii) gender-controlled \mbox{EN$\rightarrow$RO} translation. We evaluate our approach on multiple test suites including WinoMT and WinoGender \cite{winogender,winomt}, achieving significant gains in gender translation accuracy over baseline MT. Our results demonstrate that combining LLM-based context understanding with targeted NMT adaptation can mitigate gender bias in low-resource language pairs. Direct LLM translation remains costly and difficult to control at scale; our pipeline offers explicit controllability, interpretability, and compatibility with existing MT systems.

\section{Related Work}
Gender bias in machine translation (MT) has been previously documented, particularly in language pairs where the source language (e.g., English) lacks overt gender markers, while the target language (e.g., German, French, Spanish) requires grammatical gender agreement. Early studies \cite{winogender,winomt} introduced diagnostic datasets such as WinoGender and WinoMT to systematically evaluate gender bias in translation.

Recent work on gender-inclusive machine translation shows that state-of-the-art MT systems and LLMs continue to default to masculine forms, particularly in morphologically gendered languages, even when gender-neutral or gender-ambiguous translations are appropriate \cite{hackenbuchner-etal-2025-genderous,savoldi-etal-2025-mind}. Newly introduced multilingual benchmarks and evaluation datasets reveal that models struggle to exploit contextual cues, extended discourse, and explicit instructions to reliably produce inclusive or neutral forms \cite{pranav-etal-2025-glitter}. Cross-linguistic analyses further highlight persistent difficulties with gender ambiguity and non-binary constructions, motivating approaches that explicitly detect and control gender-relevant information prior to or during translation \cite{hackenbuchner-etal-2024-automatic}.

Related work has also explored sentence-level source-side gender tags, where the gender of the speaker is provided as an explicit signal to the translation model \cite{vanmassenhove2018gender}. More recently, large language models (LLMs) have been explored as an alternative to traditional tagging. Instruction-tuned models have been shown to produce gender-controlled outputs via prompt engineering \cite{sanchez-etal-2024-gender, sant-etal-2024-power}. These models can generate separate masculine and feminine translations by conditioning on contextual cues or examples. However, most of this work focuses on high-resource languages such as Spanish, French, German.

Shared-task style evaluations such as the WMT 2020 Gender Coreference and Bias task further highlight persistent gender biases across many submitted MT systems and target languages \cite{kocmi-etal-2020-gender}.

To date, gender bias in English-Romanian MT remains unaddressed. Romanian presents special challenges due to its three-gender system and complex morphological agreement. 

 \section{EnGen: The English Gender Disambiguation Dataset}
\label{sec:llm_datasets}

We build two datasets for fine-tuning LLMs to predict the gender of a target word in a given context. The process is semi-automatic - a native speaker of Romanian creates sentences, additional examples are generated by GPT through the OpenAI API~\cite{openai2024gpt4o}\footnote{Full LLM prompt is provided in the official repository.} and then the output is checked and filtered again by a native speaker.
The datasets follow a two-stage curriculum learning approach, where training starts with simpler examples and increases in difficulty: 

\subsection{Dataset 1 - Single-Entity Phrases}

\begin{figure}[ht]
  \centering
  \includegraphics[width=.33\linewidth]{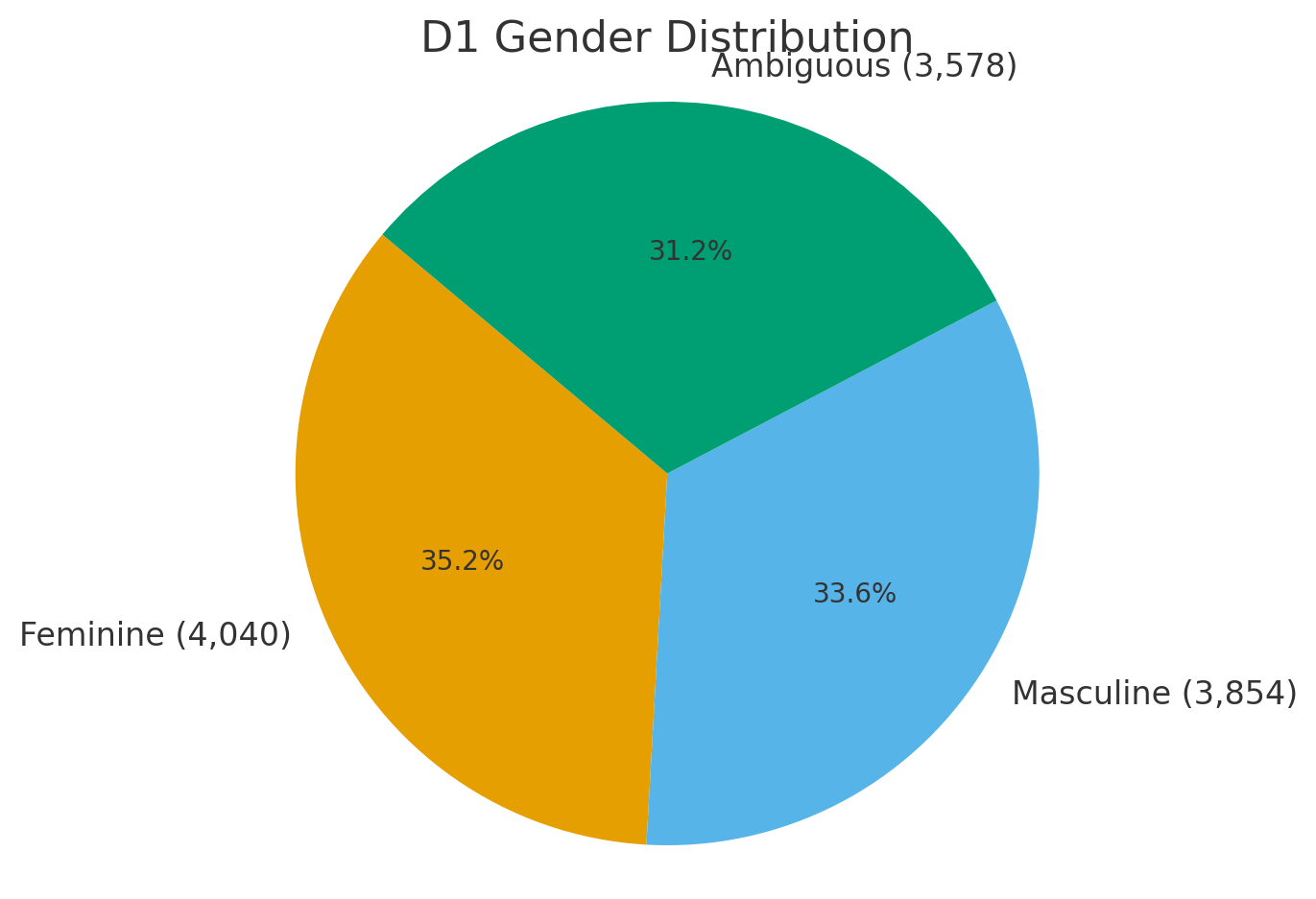}\hfill
  \includegraphics[width=.31\linewidth]{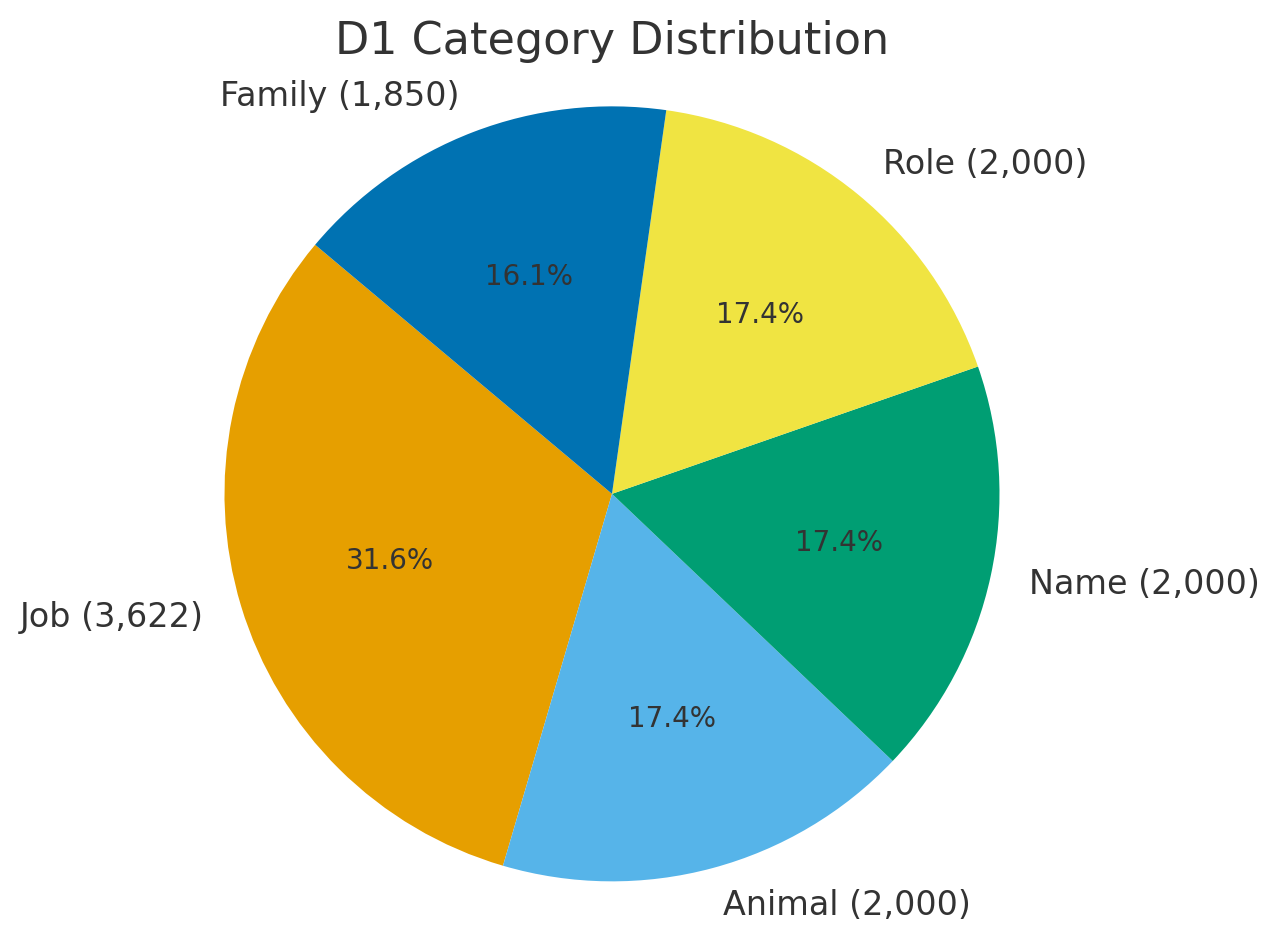}
  \includegraphics[width=.33\linewidth]{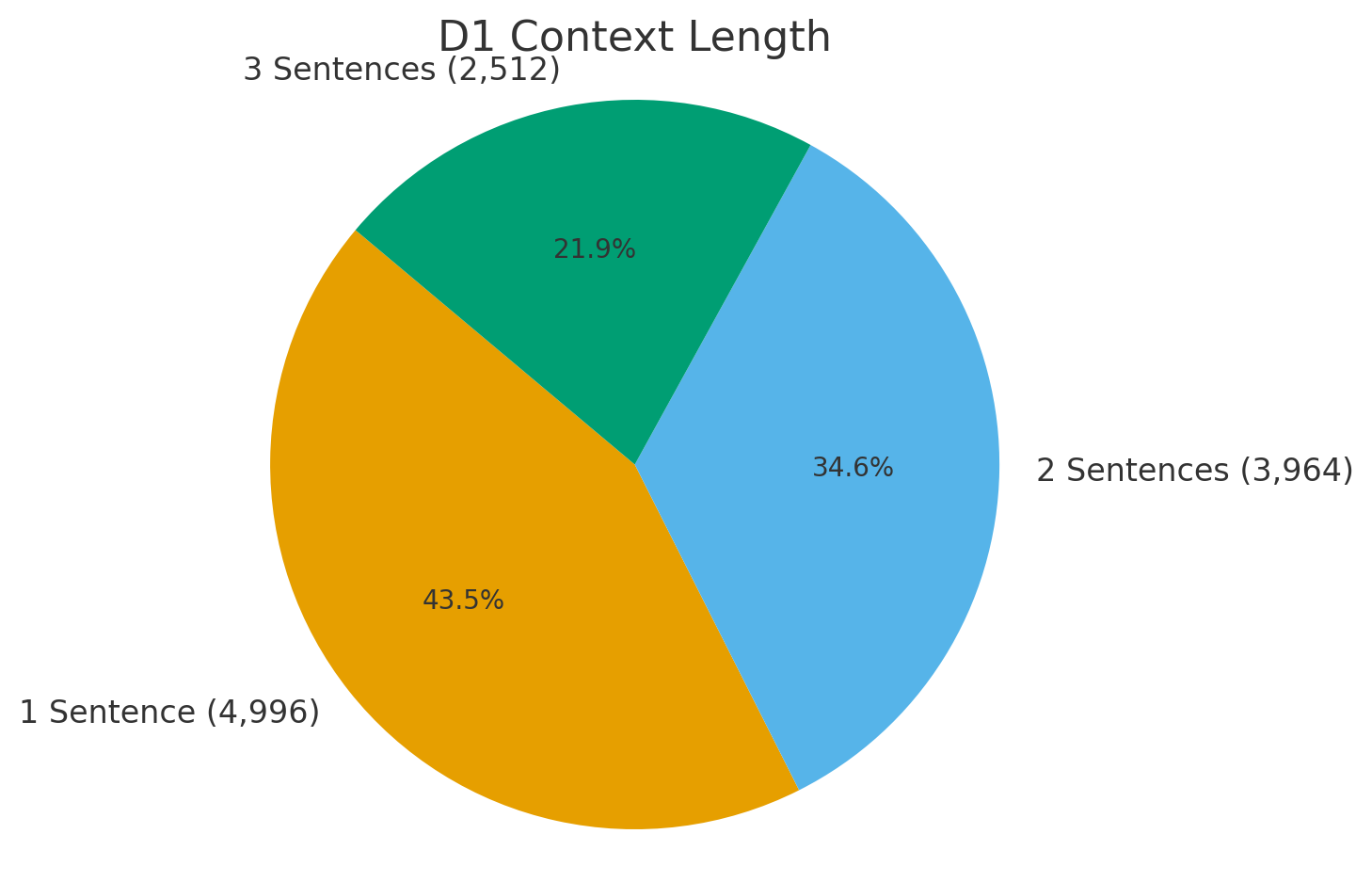}
  \caption{Dataset~1 distributions. Left: gender is equally balanced across feminine, masculine, and ambiguous entities.  Middle: category is equally balanced for nouns related to family, animals, roles, jobs, names. Right: the majority of phrases contain only one sentence, followed by two and three sentence context lengths.}
  \label{fig:d1-gender}
\end{figure}

Dataset~1 consists of 11{,}472 examples. Each example contains one genderable word, such as a job title, family member, animal, role, or proper name. The phrases may contain up to three sentences. We use ambiguous to denote cases where gender cannot be inferred from context; this category includes gender-neutral or non-binary references (e.g., ‘they’), without making assumptions about the speaker’s gender identity. Statistics regarding the distribution of each categories in the dataset are visible in \autoref{fig:d1-gender}.

The data generation process begins with one-sentence contexts. For each semantic category, a separate prompt is 
used to create examples where the gender of the target word is clear or ambiguous. Once all categories are completed for the one-sentence context, the process repeats
with two-sentence contexts, and then again with three-sentence contexts. Across all context lengths, the dataset is balanced to contain as close as possible the number of feminine, masculine, and ambiguous examples.

\begin{table}[htb]
\centering
\small
\caption{Sample examples from Dataset 1. Each sentence includes a target word and its corresponding gender.}
\label{tab:gpt_sample}
\resizebox{\textwidth}{!}{%
\begin{tabularx}{\textwidth}{@{}L{7cm}L{3cm}L{1.5cm}@{}}
\toprule
\textbf{Sentence} & \textbf{Target Word (Category)} & \textbf{Gender} \\
\midrule

The \hl{pilot} skillfully navigated the plane through turbulent weather. 
& pilot (job) & Ambiguous \\ \\

The \hl{engineer} designed an innovative solution to the problem. Her technical skills were instrumental in the project's success. The team appreciated her forward-thinking approach.
& engineer (job) & Feminine \\ \\

The \hl{leader} gave a rousing speech to the team. He inspired everyone to do their best.
& leader (role) & Masculine \\ \\

Our \hl{cousin} is an excellent cook. & cousin (family) & Ambiguous \\

\bottomrule
\end{tabularx}}
\end{table}

\subsection{Dataset 2 - Multi-Entity Phrases}
\label{sec:dataset2}

Dataset~2 contains 996 examples with unique target words.
Every sentence contains exactly two genderable words: the target word, whose gender must be predicted, and a distractor word, which serves to increase task complexity. The context length is one or two sentences. The distractor entity has ambiguous gender and is placed in various positions within the sentence to reduce predictability.

We ensure a balanced gender representation (feminine, masculine, ambiguous) while keeping the surrounding context highly similar across variants. All three gender labels appear for each target word and the context remains semantically and syntactically natural. The format is identical to dataset 1, consisting of an input (sentence, target word and category) and an output (gender label).

\begin{table}[H]
\centering
\small
\caption{Triplet Sample from Dataset 2. Here the "librarian" represents the target word and "teacher", "student", "detective" are the other entities from the context that can also be gendered.}
\label{tab:gpt4o_output_d2}
\begin{tabularx}{\textwidth}{@{}L{7cm}L{3cm}L{1.5cm}@{}}
\toprule
\textbf{Sentence} & \textbf{Target Word (Category)} & \textbf{Gender} \\
\midrule

The \hl{librarian} cataloged the new books, and the teacher borrowed a few from her.
& librarian (job) & Feminine \\ \\

The \hl{librarian} organized the shelves as the student sought guidance from him.
& librarian (job) & Masculine \\ \\

The \hl{librarian} recommended a thriller, which the detective found thrilling.
& librarian (job) & Ambiguous \\

\bottomrule
\end{tabularx}
\end{table}
\subsection{Dataset Splitting}
Both datasets are split into training, validation, and test sets to support the two-stage fine-tuning procedure, with splits designed to preserve class balance and avoid leakage.

For Dataset 1 (D1) we use an 80/10/10 split stratified by gender and category, ensuring each gender–category combination is proportionally represented. The resulting sizes are: train = 9,177, val = 1,147, test = 1,148. For Dataset 2 (D2) we use a 70/15/15 triplet-locked split: each (masculine, feminine, ambiguous) triplet receives a group ID and the whole group is assigned to exactly one partition (train, validation or test). This guarantees that no variant from the same triplet appears in different splits, including the test set. The resulting sizes are: train = 699, val = 150, test = 147. Training sets are used for parameter updates, validation sets for hyperparameter selection, and test sets are reserved strictly for final evaluation.

We adopt a curriculum learning approach: dataset~1 contains single-entity phrases so the model can first learn the basic mapping from context to gender without interference. Dataset~2 raises the difficulty with two genderable words (one target and one distractor) varying in positions and cues.

\section{EnRoGend: a Parallel English-Romanian Gender-Tagged Dataset}
\label{sec:mt_datasets}

The dataset consists of 1{,}974 examples, organized into pairs: two versions of the same sentence, each with the same target word marked for a masculine and feminine genders. Each example includes the English source, where the target word is surrounded by a gender tag <tgM>target word</tgM> or <tgF>target word</tgF>, and the corresponding Romanian translation, which has no tags. This dataset contains only occupations and person-related nouns having the purpose of teaching the MT system that it should use feminine when it sees <tgF></tgF> and masculine when it sees <tgM></tgM>.
The construction of this dataset begins with a predefined list of 82 entities referring to jobs and roles. For each entity, we write between 10 and 15 sentences where the entity's gender is elicited using the feminine pronoun (she), then we duplicate the sentence using the masculine pronoun; see for reference \autoref{tab:gender-examples}. All sentences are translated into Romanian and verified by two annotators. The dataset is balanced between gendered target words, with 987 masculine and 987 feminine instances, covering 82 distinct occupations.

\begin{table}[htb]
\centering
\small
\caption{Sample Pair from the English-Romanian Gender Tagged Dataset.}
\label{tab:gender-examples}
\begin{tabularx}{0.9\textwidth}{@{}L{0.5cm}X@{}L{3cm}}
\toprule
\textbf{En} & The teacher thanked the \hl{<tgM>lawyer</tgM>} since \hl{he} had been generous throughout the project. \\
\textbf{Ro} & Profesorul i-a mulțumit avocatului, deoarece fusese generos pe parcursul proiectului. \\
\addlinespace
\textbf{En} & The \hl{<tgF>journalist</tgF>} advised the painter on the task because \hl{she} was kind.
 \\
\textbf{Ro} & Jurnalista l-a sfătuit pe pictor pentru că era amabilă. \\
\bottomrule
\end{tabularx}
\end{table}

We split the machine–translation dataset with a similar pair-locked strategy to avoid near-duplicate leakage. Each example belongs to a 2-item minimal pair. Instead of shuffling individual rows, we shuffle pairs and keep both members together in the same partition. The splits are 80\% train, 10\% validation and 10\% test.
This guarantees that if the masculine variant of a sentence is in training, its feminine counterpart cannot appear in validation or test (and vice versa), including for the held-out test set. The result is a fair evaluation that is not inflated by near-duplicate examples.

\section{Methodology}
\begin{figure}[!htb]
\centering
\scriptsize
\begin{tikzpicture}[
    node distance=0.3cm,
    >=latex,
    every node/.style={align=center},
    box/.style={draw, rounded corners, fill=blue!7, minimum width=4cm, minimum height=0.5cm},
    arrow/.style={->, thick}
]

\node[box] (input) {\textbf{User Input Sentence} \\ 
The doctor congratulated the nurse because he did a good job.};

\node[box, below=of input] (entities) {\textbf{Entity Selector} \\ 
Identifies: doctor, nurse};

\node[box, below=of entities] (llm) {\textbf{LLM Gender Classification} \\ 
\begin{tabular}{l l}
Target word: doctor $\rightarrow$ ambiguous \\
Target word: nurse $\rightarrow$ masculine
\end{tabular}};

\node[box, below=of llm] (tagged) {\textbf{Merged Tagged Sentence} \\
The doctor congratulated the \hl{<tgM>nurse</tgM>} because he did a good job.};

\node[box, below=of tagged] (mt) {\textbf{Machine Translation (MT)} \\ 
Output: Doctorul l-a felicitat pe \hl{asistent} pentru că a făcut o treabă bună.};

\draw[arrow] (input) -- (entities);
\draw[arrow] (entities) -- (llm);
\draw[arrow] (llm) -- (tagged);
\draw[arrow] (tagged) -- (mt);

\end{tikzpicture}

\caption{
End-to-end pipeline example. The system takes an English sentence, extracts candidate entities, classifies gender for each using the LLM, inserts tags for gendered entities, and translates the tagged sentence into Romanian.
}
\label{fig:full_pipeline}
\end{figure}
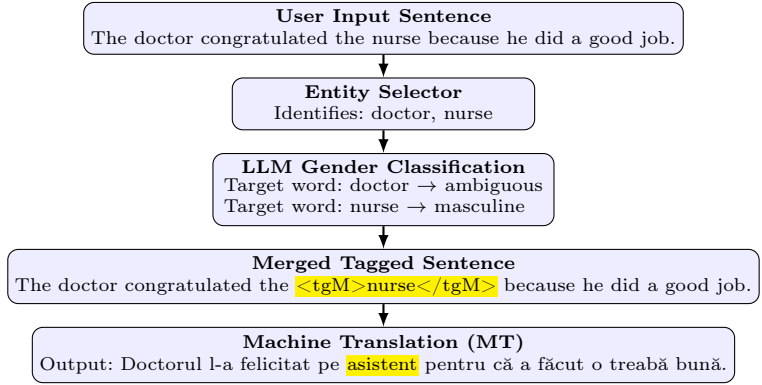

The datasets described in previous sections are the first ones to address gender bias for English-Romanian language pairs and can thus enable the creation of an end-to-end machine translation pipeline.
The pipeline processes an input sentence through a sequence of analysis and generation stages designed to preserve intended gender information during translation. Candidate entities that may require gendered realization are first identified, after which a large language model infers the contextual gender of each entity based on discourse cues. The inferred gender information is then encoded using inline tags and merged back into the original sentence, forming an intermediate representation that makes gender constraints explicit. This tagged sentence is subsequently passed to the machine translation system, which uses the annotations to guide the selection of appropriate gendered forms in the target language, thereby reducing reliance on default or biased gender choices. The entire process is rendered in \autoref{fig:full_pipeline}.

To enable gender-aware processing in realistic settings, the pipeline relies on an \textit{Entity Selector} that used a predefined list of gendered entities that serve as candidate targets for gender inference. The list is compiled based on previous studies \cite{entitylist}. Each word in the input sentence is matched against this list to identify role nouns or occupations whose gender must be inferred. To improve robustness, fuzzy string matching is applied to account for misspellings, plural forms, and minor lexical variations.

\subsection{LLM Gender Classification}
For this step, the main focus is to have a classifier that is fast and easily deployable in an MT system. As such, we employ the pre-trained LLaMA~3.2 (1B parameters) model from Unsloth \cite{unsloth_llama32_1b_4bit}. To make fine-tuning feasible on limited hardware, the model is loaded in 
4-bit \texttt{NF4} quantization and fine-tuned with mixed precision.  
We adopt Low-Rank Adaptation (LoRA) \cite{hu2022lora}.  
Instead of updating a full weight matrix $W_0 \in \mathbb{R}^{d \times k}$, LoRA learns a low-rank update:
\[
W = W_0 + \Delta W, \qquad 
\Delta W = \tfrac{\alpha}{r} AB,
\]
where $A \in \mathbb{R}^{d \times r}$, $B \in \mathbb{R}^{r \times k}$, $r \ll \min(d,k)$, 
and $\alpha$ scales the update.  
The frozen $W_0$ preserves pre-trained knowledge, while $\Delta W$ captures task-specific adjustments.  
LoRA adapters are inserted into the query, key, value, output, gate, up, and down projection layers.  
Only LoRA parameters and the classification head are updated during training. We use AdamW with separate parameter groups, one for the LoRA adapter parameters and one for the language modeling head. This allows assigning distinct learning rates to each group, while weight decay is applied only where specified in the grid configuration. Training is performed with gradient accumulation and mixed-precision computation in \texttt{bfloat16}, on a single Nvidia L4 GPU in Colab Pro.

\subsubsection*{Classification Strategy} a Linear Classification Head is attached to the pre-trained LLaMA~3.2 (1B parameters ) model. After processing the input sequence through the language model, we extract the hidden state corresponding to the final token position in the sequence. This hidden representation, denoted $\mathbf{h}_{\text{last}}$, encodes the full context of the input, including the sentence and target word.  

A fully connected linear layer then transforms this hidden vector into a fixed-size output vector representing class logits for the three possible gender labels: feminine, masculine, and ambiguous. Formally, the prediction is computed as:
\[
\mathbf{h}_{\text{last}} = \text{LLM}(\mathbf{x})[-1], \qquad
\mathbf{y} = \mathbf{W}\mathbf{h}_{\text{last}} + \mathbf{b},
\]
where $\mathbf{W} \in \mathbb{R}^{3 \times d}$ and $\mathbf{b} \in \mathbb{R}^3$ are the learned parameters of the classification head, and $d$ is the dimensionality of the hidden state. Model training is guided by the cross-entropy loss.

\subsubsection{The Two-Stage Curriculum Learning}
\label{sec:two_stage}
is designed to gradually increase task complexity while training. In the first stage, the model is fine-tuned on Dataset~1, which contains sentences with a single named entity and well-balanced gender classes. These examples provide both clear and ambiguous contexts, allowing the model to learn the core gender classification task with minimal distraction or noise. 

Once this base ability is acquired, the second stage uses the Dataset~2, which presents more realistic challenges, including sentences containing multiple named entities. This stage encourages the model to reason about context and resolve gender cues in more complex linguistic scenarios.

\subsubsection*{Hyperparameter Search and Evaluation}
\label{sec:hparam_search}

We explore hyperparameter configurations for both stages of curriculum training:

\begin{itemize}
    \item Stage 1 (initial fine-tuning on D1): LoRA rank $r \in \{8, 12, 16, 24, 32\}$ with scaling $\alpha \approx 2r$, dropout in $[0.00, 0.06]$, learning rates for the LoRA adapters and classifier head ($4\times10^{-5}$--$1.6\times10^{-4}$), weight decay in $\{0.0, 0.005, 0.01\}$, and $2$--$3$ epochs.
    \item Stage 2 (continued fine-tuning on D2): LoRA ranks $\{8, 12, 16\}$ with proportionally scaled $\alpha$, dropout in $[0.04, 0.07]$, smaller learning rates ($4\times10^{-5}$--$1.0\times10^{-4}$), weight decay in $\{0.005, 0.01\}$, and $1$--$2$ epochs.
\end{itemize}

For hyperparameter tuning we use optuna search. It runs a fixed budget of trials (e.g., 12), guided by a Tree-structured Parzen Estimator (TPE) sampler.

We observe that several configurations achieve high accuracy and F1 scores on Dataset~1, with the top-performing model reaching an F1 of 0.97 and accuracy of 97\% on the test set.
The performance on Dataset~2 varies more widely due to its complex, multi-entity structure. While the best model achieves a strong F1 score of 0.95, others drop as low as 0.66.
Overall, the performance is sensitive to regularization and learning rate balance, and improves with moderate-to-deep LoRA ranks (e.g., 12–16) paired with well-scaled $\alpha$ values.

Table~\ref{tab:joint} summarizes the generalization in both datasets. The best joint test F1 score reaches 0.96, confirming that high-quality performance can be maintained by generalizing across D1 and D2 simultaneously. 

In addition, we evaluate the final Stage~2 models both on D2 and again on D1, which allows us to quantify catastrophic forgetting via the Forgetting F1 metric (see \autoref{tab:joint}). This approach makes the runs easy to compare and gives us a reliable way to measure the effect of curriculum learning. The joint metrics represent the model's performance on the combined test set, which merges the test splits from both datasets, D1 and D2. The Forgetting F1 metric shows how much performance on D1 is lost after the second stage of training on D2. It is computed as the difference in F1 score on the D1 test set after and before Stage~2 fine-tuning, a negative value indicates forgetting (performance decreased), while a positive value implies improvement or recovery on D1 after further training.

\begin{table}[htb]
\centering
\caption{The Forgetting F1 metric quantifies the degree of catastrophic forgetting, with most configurations showing mild drops (e.g., -0.0121 or -0.0015), while a few show large negative values (e.g., -0.1820), indicating significant loss of earlier knowledge.
}
\label{tab:joint}
\begin{tabular}{llrrrr}
\toprule
\textbf{D2 idx} & \textbf{D1 idx} &
\textbf{Joint Acc} & \textbf{Joint F1} & \textbf{Joint MCC} &
\textbf{Forgetting  F1} \\
\midrule
\textbf{D2-1} & \textbf{D1-1} &
\textbf{96\%} & \textbf{0.96} & \textbf{0.94} & \textbf{0.0083} \\

D2-2  & D1-2 & 83\% & 0.82 & 0.75 & -0.1040 \\
D2-3  & D1-3 & 95\% & 0.95 & 0.93 & -0.0121 \\
D2-4  & D1-1 & 96\% & 0.96 & 0.94 &  0.0072 \\
D2-6  & D1-2 & 80\% & 0.80 & 0.70 & -0.0851 \\
D2-8  & D1-3 & 95\% & 0.95 & 0.93 & -0.0015 \\
D2-9  & D1-2 & 73\% & 0.74 & 0.60 & -0.0851 \\
D2-10 & D1-4 & 86\% & 0.86 & 0.81 & -0.1820 \\
D2-11 & D1-5 & 66\% & 0.66 & 0.58 &  0.1950 \\
\bottomrule
\end{tabular}

\end{table}

\subsection{Gender-Aware Machine Translation (En\texorpdfstring{$\rightarrow$}{→}Ro)}
\label{sec:mt_pipeline}

We fine-tune an English-to-Romanian pre-trained Transformer model Helsinki-NLP/opus-mt-en-ro~\cite{opusmt_en_ro} 
and evaluate performance on both in-domain splits and external diagnostic sets. The tokenizer is extended with four special tokens: <tgM>, </tgM>, <tgF>, and </tgF> on the source side. If no padding token is defined, the EOS token is reused as PAD. Romanian references are kept tag-free to avoid leaking gender labels into the target side.

Because the dataset contains many minimal pairs 
(i.e., identical English sentences differing only in masculine vs.\ feminine tags), 
a random split would risk leaking near-duplicate examples across train, validation, and test sets, leading to artificially inflated evaluation scores. 
To prevent this, we enforce a pair-locked split:

\begin{itemize}
    \item Sentences are grouped into pairs consecutively by file order: \((0,1),(2,3),\dots\). 
          Each pair is assigned a unique integer pair\_id defined as \(\lfloor i/2 \rfloor\). 
          We ensure an even number of rows; otherwise, the source file is adjusted.
    \item Exact duplicates on the (English, Romanian) tuple are removed to avoid trivial matches at evaluation time.
    \item The set of unique pair\_ids is randomly shuffled with a fixed seed for reproducibility.
    \item The first 80\% of pair\_ids are assigned to the training set, 
          the next 10\% to validation, and the remaining 10\% to test. 
          Both members of each minimal pair are always placed in the same split.
    \item We verify that pair\_id sets are disjoint across splits and 
          report hashed overlaps on both source and target texts as a 
          sanity check against accidental data leakage.
\end{itemize}

\subsubsection*{Hyperparameter Tuning}
\label{sec:mt_hparam}
To determine the most effective fine-tuning strategy for Transformer with gender tags, 
we explore three adaptation regimes under identical training loops:
\begin{itemize}
\item Full fine-tuning: all model parameters are updated, providing maximum flexibility but at a higher computational cost.  
\item Partial fine-tuning: most encoder layers are frozen, and only the decoder, shared embeddings, language modeling head, and the last \(N\) encoder layers (\(N \in \{1,2,3\}\)) are updated, reducing training cost while retaining adaptability.  
\item LoRA fine-tuning: a parameter-efficient strategy where low-rank adapters are injected into the attention layers, keeping the base model frozen and updating only the adapter parameters.
\end{itemize}
Hyperparameter tuning is performed using Optuna with a grid-based search over the following space: learning rate 
$\in \{1\mathrm{e}{-5}, 5\mathrm{e}{-5}, 1\mathrm{e}{-4}\}$, batch size $\in\{8,16\}$, epochs $\in\{1,2\}$, and, for partial fine-tuning, 
the number of unfrozen encoder layers $N_{\text{unfreeze}} \in\{1,2,3\}$.  
Model selection is based on validation performance, and the best configurations are evaluated on the held-out test set using BLEU \cite{post-2018-call}, chrF++ \cite{popovicchr}\footnote{chrF++ signature is \path{nrefs:1|case:mixed|eff:yes|nc:6|nw:0|space:no|version:2.4.3} and BLEU signature is \path{nrefs:1|case:mixed|eff:no|tok:13a|smooth:exp|version:2.4.3}.}, TER \cite{snover-etal-2006-study}, and COMET \cite{cometdaxl} metrics. The reported metrics reflect different aspects of translation quality: BLEU and chrF++ measure word and character n-gram overlap between system output and reference; TER reflects the number of edits needed to reach the reference, where lower is better; and COMET is a learned metric that evaluates adequacy and fluency based on source, hypothesis, and references.
The surface-form metrics such as BLEU, chrf++, and TER tend to reflect good results due to the high overlap, however the gender mismatch is more nuanced semantic aspect, therefore we rely more on the COMET metric for the final judgments and keep the remaining metrics as evidence.

\begin{table}[htb]
\centering
\caption{
Hyperparameter optimization results on the \textbf{validation set} for fine-tuning Transformer with gender tags.
The table compares full, partial, and LoRA fine-tuning modes across various learning rates, batch sizes, epochs,
and (for partial mode) number of unfrozen encoder layers.
Full fine-tuning with a learning rate of $1\mathrm{e}{-4}$, batch size 16,
and 2 epochs achieved the best validation performance, with the highest BLEU (97.29), chrF (98.57), COMET (0.853),
and the lowest TER (1.71).
} 
\label{tab:mt_optuna}
\begin{tabular}{lcccccccc}
\toprule
Mode & {LR} & {Batch} & {Epochs} & {$N_{\text{unfreeze}}$} &
{BLEU} & {chrF++} & {TER} & {COMET} \\
\midrule
partial & 1.00e-05 & 16 & 2 & 1 & 63.69 & 80.31 & 24.88 & 0.7723 \\
full    & 5.00e-05 &  8 & 1 & {} & 93.03 & 96.73 &  3.55 & 0.8409 \\
partial & 1.00e-04 & 16 & 1 & 1 & 92.94 & 96.40 &  3.90 & 0.8396 \\
full    & 1.00e-05 &  8 & 1 & {} & 67.57 & 83.28 & 21.50 & 0.7857 \\
lora    & 1.00e-04 &  8 & 1 & {} & 41.12 & 76.78 & 44.84 & 0.7263 \\
\textbf{full} & \textbf{1.00e-04} & \textbf{16} & \textbf{2} & {} &
\textbf{97.29} & \textbf{98.57} & \textbf{1.71} & \textbf{0.8530} \\
lora    & 1.00e-04 & 16 & 2 & {} & 41.71 & 67.35 & 43.98 & 0.7261 \\
lora    & 1.00e-05 &  8 & 1 & {} & 23.99 & 57.59 & 71.13 & 0.6467 \\
lora    & 5.00e-05 &  8 & 1 & {} & 33.08 & 61.98 & 53.70 & 0.7103 \\
partial & 1.00e-05 & 16 & 2 & 3 & 69.30 & 84.06 & 20.13 & 0.7853 \\
partial & 5.00e-05 & 16 & 1 & 3 & 89.64 & 94.95 &  5.61 & 0.8306 \\
partial* & 1.00e-04 &  8 & 1 & 2 & 97.07 & 98.40 &  1.93 & 0.8511 \\
\bottomrule
\end{tabular}

\end{table}

The results in Table~\ref{tab:mt_optuna} show that full fine-tuning with a learning rate of $1\mathrm{e}{-4}$, 
batch size 16, and 2 epochs achieved the best overall performance (highlighted in boldface). Partial fine-tuning with 2–3 unfrozen encoder layers performed competitively (bottom row marked with an *), achieving a COMET scores above 0.85 and comparable overall scores to the full-finetuning model. This represents a good compromise between efficiency and accuracy.  

The best hyperparameter configuration selected on the validation set based on the results of \autoref{tab:mt_optuna} is the fully finetuned model. This model achieves similar scores on the held-out test set: a COMET score of 0.846, a BLEU score of 96.92, a chrF++ score of 98.21, and a TER of 1.94.

LoRA, however, underperformed in our experimental setting, suggesting that lightweight adapter tuning 
does not provide enough capacity for this task. A likely explanation is that this task requires more than light domain adaptation. The model must learn to use new source-side gender tags and realize them through correct Romanian morphology, including agreement on nouns, adjectives, and verbs. Full and partial fine-tuning are better suited to this controlled generation setting because they let more of the model adapt to the tag signal. By contrast, the LoRA setups tested may have been too limited to capture the link between explicit gender tags and downstream morphological realization. The large gap in performance therefore suggests that this form of parameter-efficient tuning is not sufficient for robust gender control in English–Romanian translation. Another possible factor is that Romanian morphological agreement is more demanding than the target-side changes required in some higher-resource settings.

\section{Benchmark Evaluation}

To evaluate gender bias in machine translation, we rely on several benchmarks that are designed to probe whether a system can correctly resolve gender based on cues and whether it reflects correct grammatical agreement in gender-marked target languages.
Although GPT-4o was used in the data creation process, all benchmark evaluations are conducted on external datasets that are not used during training, to ensure that there are no data leaks. In addition, we have compared all the sentences with our dataset to ensure that there is no data leakage.

\paragraph{The WinoMT dataset~\cite{winomt}} consists of (1,584 pro and 1,584 anti) English sentences in which gender must be inferred from context. Each sentence is presented in two forms: a \textbf{pro-stereotypical} variant, where the pronoun aligns with common gender stereotypes  and an \textbf{anti-stereotypical} variant, where the pronoun contradicts such stereotypes, e.g., pro-stereotypical sentence: \textit{The nurse helped the patient because \hl{she} was kind.} Anti-stereotypical sentence: \textit{The nurse helped the patient because \hl{he} was kind.} 
An unbiased system is expected to perform similarly on both pro-stereotypical and anti-stereotypical examples. 

\paragraph{The WinoGender dataset~\cite{winogender}} is a pronoun resolution benchmark that tests the impact of gendered pronouns on translation, consisting of 720 sentences. Each example is a minimal pair differing only in the pronoun (he, she or they) and is used to decide whether the system’s output reflects these distinctions correctly in the gendered target-language translation. 

All English test sentences are translated using three systems: 1. Raw MT - the base Transformer model, without any additional gender hint tags; and 2. Gender-Aware Pipeline - each sentence passed through our pipeline (as described in \autoref{fig:full_pipeline}); 3. GPT-5.2 using default system settings and zero-shot translation prompts; due to its proprietary nature, exact replication may not be possible. These test sets do not include gold-standard Romanian reference translations. Therefore, a manual evaluation was conducted by a native Romanian speaker and verified through spot checks. For each test case, we check whether the translation preserves the correct grammatical gender of the target word, along with agreement (e.g., adjective inflection, verb conjugation).

\begin{table}[htb]
\centering
\caption{The results presented in the table are the accuracies on benchmark test sets. It shows that the raw Transformer system performs better when the correct gender is masculine, as it often defaults to masculine forms during translation. While this leads to higher scores on Pro-stereotypical examples (where the gold label aligns with masculine bias), it harms performance on Anti-stereotypical and feminine cases. Similarly, a state-of-the-art model such as GPT-5.2 has a strong preference for stereotypical biases.
The gender-aware pipeline significantly improves accuracy across all subsets by explicitly guiding the model toward the intended gender.}
\label{tab:mt_gender_bias}
\begin{tabular}{lccc}
\toprule
\textbf{Model / Metric} & \textbf{WinoMT (Pro)} & \textbf{WinoMT (Anti)} & \textbf{WinoGender} \\
\midrule
\textbf{Raw Transformer} & 59.05\% & 50.13\% & 50.28\% \\
\textbf{Pipeline LLM-MT} & \textbf{96.34\%} & \textbf{93.68\%} & \textbf{91.53\%} \\
\textbf{GPT-5.2} & 79.23\% & 58.79\% & 67.71\% \\
\bottomrule
\end{tabular}
\end{table}

Some translations contain issues such as incorrect word choices or missing diacritics. However, as these errors occur in both raw and fine-tuned outputs, we consider a translation correct if the intended gender is correct and if it stays in agreement with the rest of the sentence.
For example, if the MT translates salesperson for feminine gender as "vânzătora" instead of "vânzătoarea" we consider it correct.

\section{Conclusion}
We investigate gender bias in English$\rightarrow$Romanian machine translation, where ambiguous English inputs must be rendered with explicit grammatical gender in Romanian. Our results confirm that a standard Transformer MT baseline frequently defaults to masculine forms and exhibits stereotype sensitivity, performing substantially better on pro-stereotypical WinoMT examples than on anti-stereotypical ones. Furthermore, even proprietary state-of-the-art models such as GPT-5.2 have a bias towards stereotypical translations into Romanian, despite the fact that such models might have been exposed to the WinoMT and WinoGender datasets.

To mitigate gender bias, we propose a hybrid pipeline that combines LLM-based contextual gender disambiguation with tag-aware neural machine translation: a fine-tuned LLM predicts the intended gender of target entities in the English source and inserts inline gender hint tags, which a Transformer model learns to follow during translation. Across external diagnostic benchmarks, this approach yields large gains in gender correctness, improving accuracy by over 40 percentage points relative to the raw MT system and substantially reducing the pro/anti performance gap, indicating reduced reliance on stereotypical defaults.

To support research in this low-resource setting, we introduced several datasets for (i) EnGen - English language gender disambiguation with a curriculum learning setup and (ii) EnRoGend - a controlled En$\rightarrow$Ro translation with gender tags, using leakage-safe splitting strategies. We also found that full and partial fine-tuning of the MT model effectively leverage the gender tags, while the LoRA configurations we tested underperformed for this task, suggesting that tag-conditioned morphological control may require greater adaptation capacity.

While our experiments focus on Romanian, the proposed pipeline is applicable to other morphologically gendered target languages.

\section*{Acknowledgments}
This research is supported by InstRead: Research Instruments for the Text Complexity, Simplification and Readability Assessment  CNCS - UEFISCDI project number PN-IV-P2-2.1-TE-2023-2007 and by the project ``Romanian Hub for Artificial Intelligence - HRIA'', Smart Growth, Digitization and Financial Instruments Program, 2021-2027, MySMIS no. 351416.

\section*{Note}
This preprint has not undergone peer review or any post-submission improvements or corrections. The Version of Record of this contribution is published in Lecture Notes in Computer Science (LNCS, Springer), and is available online at
\url{https://doi.org/10.1007/978-3-032-29532-3_11}

%
%
\bibliographystyle{splncs04}
\begingroup
\let\clearpage\relax
\bibliography{custom}
\endgroup

\end{document}